\documentclass[10pt,twocolumn,letterpaper]{article}

\usepackage{cvpr}
\usepackage{times}
\usepackage{epsfig}
\usepackage{graphicx}
\usepackage{amsmath}
\usepackage{amssymb}
\usepackage{multirow}
\usepackage{pifont}

\newcommand{\yang}[1]{\textcolor{black}{#1}}
\newcommand{\wan}[1]{\textcolor{black}{#1}}

\cvprfinalcopy 

\def\cvprPaperID{59} 

\ifcvprfinal\fi

\begin{document}

\title{When Person Re-identification Meets Changing Clothes}

\author{Fangbin Wan$^{1*\ddagger}$\quad Yang Wu$^{2*\P}$\quad Xuelin Qian$^{3\ddagger}$\quad Yixiong Chen$^{1\ddagger}$\quad Yanwei Fu$^{1\dagger\mathsection}$\\
\\
$^1$School of Data Science, Fudan University ~~~~   $^2$Kyoto University \\ $^3$School of Computer Science, Fudan University}

\maketitle
\thispagestyle{empty}

\renewcommand{\thefootnote}%
{\fnsymbol{footnote}}
\footnotetext[1]{indicates equal contributions.} 
\footnotetext[2]{indicates corresponding author.}
\footnotetext[3]{Fangbin Wan, Xuelin Qian and Yixiong Chen are with Shanghai Key Lab of Intelligent Information Processing, Fudan University. Email: \{fbwan18, xlqian15, 16307110231\}@fudan.edu.cn.}
\footnotetext[4]{Yanwei Fu is with MOE Frontiers Center for Brain Science, and Shanghai Key Lab of Intelligent Information Processing, Fudan University. Email: yanweifu@fudan.edu.cn.}
\footnotetext[5]{Yang Wu is with Department of Intelligence Science and Technology, Graduate School of Informatics, Kyoto University. Email: wu.yang.8c@kyoto-u.ac.jp.}

\maketitle

\begin{abstract}
Person re-identification (\yang{ReID}) is now an active research topic for AI-based video surveillance applications such as specific person search, but the practical issue that the target person(s) may change clothes (clothes inconsistency problem) has been overlooked for long. For the first time, this paper systematically studies this problem. We first overcome the difficulty of lack of suitable dataset, by collecting a small yet representative real dataset for testing whilst building a large realistic synthetic dataset for training and deeper studies. Facilitated by our new datasets, we are able to conduct various interesting new experiments for studying the influence of clothes inconsistency. We find that changing clothes makes \yang{ReID} a much harder problem in the sense of bringing difficulties to learning effective representations and also challenges the generalization ability of previous \yang{ReID} models to identify persons with unseen (new) clothes. Representative existing \yang{ReID} models are adopted to show informative results on such a challenging setting, and we also provide some preliminary efforts on improving the robustness of existing models on handling the clothes inconsistency issue in the data. We believe that this study can be inspiring and helpful for encouraging more researches in this direction. The dataset is \wan{available} on the project website: https://wanfb.github.io/dataset.html.

\end{abstract}

\section{Introduction}

Automatically searching for a specific person (e.g., a suspect) across multiple video surveillance cameras at different places is a popular imagination in AI-based science fiction movies and TV dramas. It is not only extremely valuable and impactful for public safety and security (e.g., finding the Boston Marathon bombing suspect \wan{efficiently}), but also technically highly challenging. \yang{The key component of this task is referred to as person re-identification (ReID) in the literature.} It is about fine-grained search over numerous similar-looking candidates while at the same time having to tolerate significant appearance changes of the same targets.

\begin{figure}
\begin{centering}
\includegraphics[width=0.85\columnwidth]{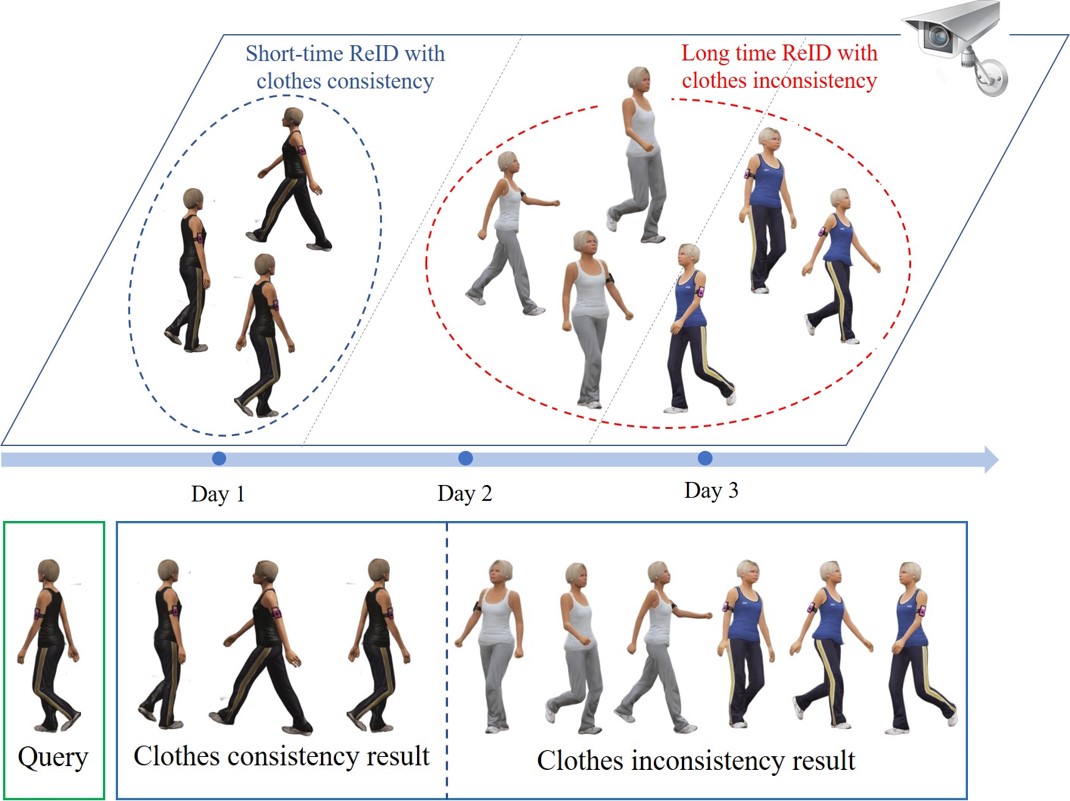} 
\par\end{centering}
\caption{\label{fig:new_task} The appearance of the same person across different days. There usually exists clothes consistency \yang{in} a short \yang{time-span}, while in a longer \yang{period}, the person \yang{usually} changes \yang{his/her} clothes. Given a query image, \yang{ReID shall} target to find both the clothes consistency and clothes inconsistency results.}
\end{figure}

The identity of a person is usually determined by his/her biological traits/characteristics, such as facial appearance, ages, body shape and so on, rather than \wan{the outlook} appearance like hairstyles, bags, shoes, and clothes \emph{et al}. In contrast, the widely used benchmarks in ReID -- Market1501 \cite{zheng2015scalable}, CUHK03 \cite{li2014deepreid}, DukeMTMC-reid \cite{zheng2017unlabeled}) do not really contain significant appearance variance of the same person, which can easily mislead ReID models to learn appearance features (\emph{e.g.}, clothes) rather than the desired robust identity-sensitive features related to biological traits. This is particularly true, in the sense that, most successful DNN-based ReID models are learning from data. The clothes consistency on the same identities makes the models greatly rely on the visual appearance related to clothes, which usually takes the largest part of a human body. Thus, these models can easily fail when people wear similar clothes as others' or change their own clothes.

Though people do change clothes frequently, especially during different days, and crime suspects may even intentionally do that on the same day for more effective hiding, the \textbf{clothes inconsistency} problem has long been overlooked in the research community. There are mainly two reasons for that. Firstly, a traditional assumption about ReID is that it only covers a short period of a few minutes/hours, so that there only exists \textbf{clothes consistency} for the same identities. However, even in such a short period, one can still take on or take off his/her coat, grasp a bag, or get dressed in another suit, let alone in long periods like across days. This assumption greatly limits the generalization ability of ReID models towards real-world applications, such as finding crime suspects. To this end, it is imperative to investigate the ReID models in the settings that allow clothes inconsistency, as shown in Fig. \ref{fig:new_task}. Secondly, it is very expensive to create a large-scale person ReID dataset with \wan{a} significant amount of same-identity clothes inconsistency in real scenarios. Capturing person image or video data may not be \yang{as} easy as before, since surveillance data is now under strict control and individual privacy is a highly valued issue \yang{in many countries}. Besides, associating the same identities with clothes changes in such large amounts of data for getting ground-truth labels is even more challenging for human annotators. These difficulties discourage the study of person \yang{ReID} with changing clothes to some extent. However, there are still very few efforts on this direction, \textit{e.g.} RGB-D dataset \cite{barbosa2012re} and the PRCC dataset \cite{yang2019person}. Although these datasets enable some preliminary research of the clothes inconsistency issue in person \yang{ReID, they} are quite limited in both the size and collected environment.

\begin{figure*}
\begin{centering}
\includegraphics[width=0.8\textwidth]{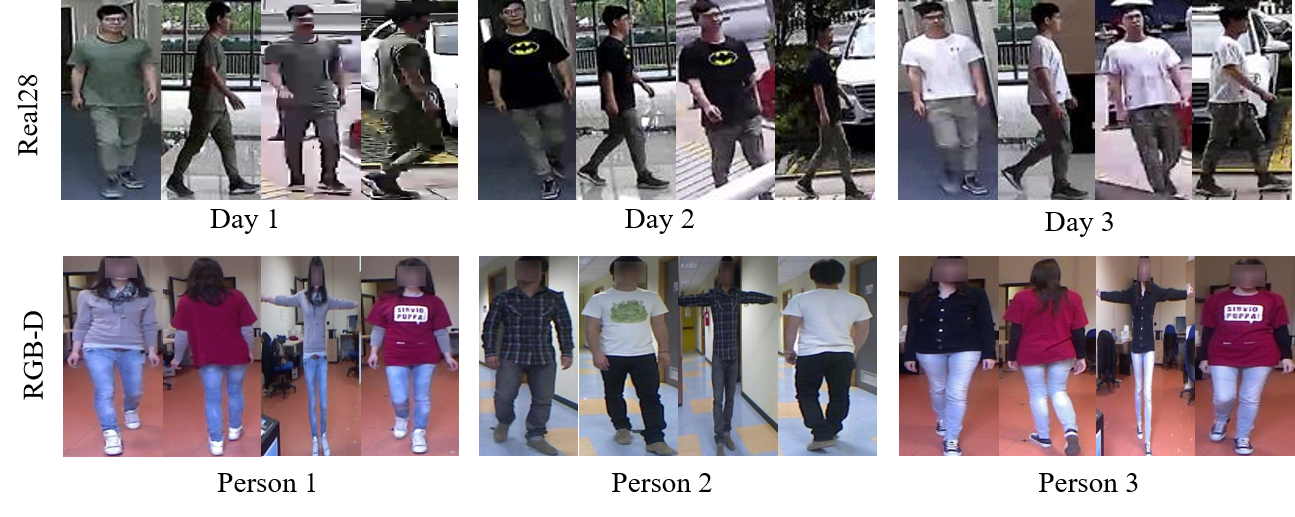} 
\par\end{centering}
 \caption{\label{fig:real_samples} { Some samples of ``Real28''
collected in different days of the same person and ``RGB-D'' captured in different days but grouped by different identity of persons.}}
\end{figure*}

\begin{figure}
\begin{centering}
\includegraphics[width=0.8\columnwidth]{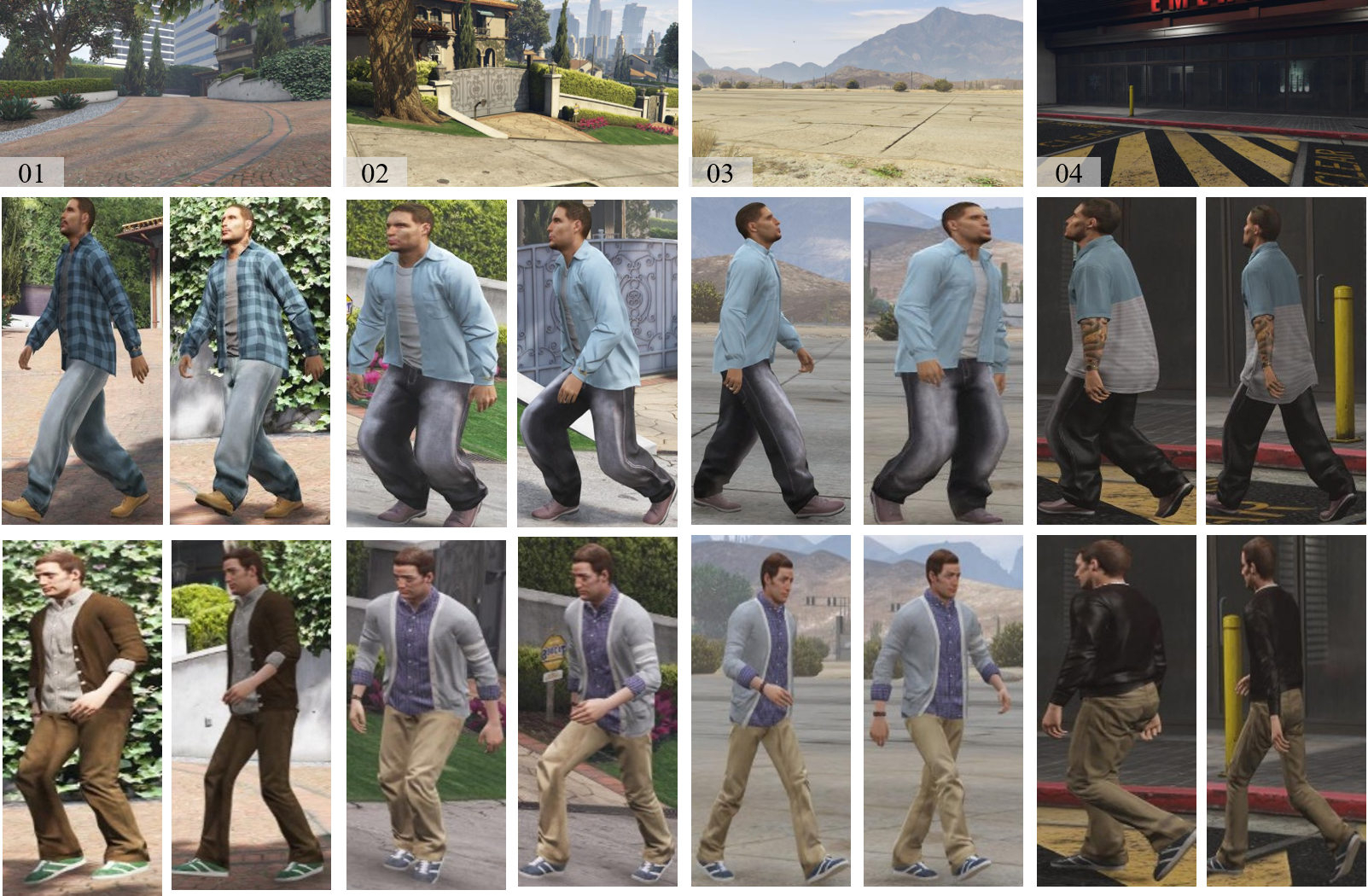} 
\par\end{centering}
\caption{Samples of VC-Clothes. The first row is the four environment maps,
the following two rows are the same persons under the 4 cameras
with at most 3 suits of clothes.} \label{fig:VC-Clothes_scene}
\end{figure}

Despite \wan{these} difficulties, in this paper, we provide benchmark datasets and systematically study the influence of clothes inconsistency problem for person ReID, which is the first time as far as we know. In more details, two benchmark datasets are built and released. One is a 3-day 4-camera dataset capturing real persons and scenarios around the entrance of a big office building (as shown in the upper row of Fig. \ref{fig:real_samples}). Due to the privacy concerns, currently we have only been able to capture data of 28 volunteers. Such difficulties motivated us to build another dataset -- a synthetic dataset called \textbf{V}irtually \textbf{C}hanging-\textbf{Clothes} (\textbf{VC-Clothes}) dataset using 3D human models with the help of a video game engine named GTA5 (samples shown in Fig.~\ref{fig:VC-Clothes_scene}). Thanks to the encouraging breakthroughs of computer graphical techniques and the prevalence of virtual reality \wan{games}, we could get high-quality realistic data covering 512 identities of 19,060 images in 4 different scenes with significant clothes changes.

The clothes inconsistency problem \yang{makes} person \yang{ReID} much more challenging\yang{, which can be easily imagined. However},  it is never validated and discussed experimentally in previous studies. In Sec. \ref{sec:Influence}, we show that clothes changes not only significantly influence representation learning, but also challenge the generalization ability of ReID models in re-identifying the persons wearing unseen clothes. Moreover, we present some preliminary efforts on how to promote performance by learning more identity-sensitive and clothes-insensitive features, which can serve as baselines for future deeper studies.

\vspace{0.1in}
\textbf{Contributions}. (1) We systematically study the influence of \wan{the} same-identity clothes inconsistency issue for ReID, which is historically overlooked. (2) We build two new benchmark datasets, one real and one synthetic with significant clothes changes, for supporting the research. (3) We provide preliminary solutions for handling the clothes inconsistency issue, which can be baselines for inspiring further researches.

\begin{figure*}[t]
\centering{}\includegraphics[width=1.8\columnwidth]{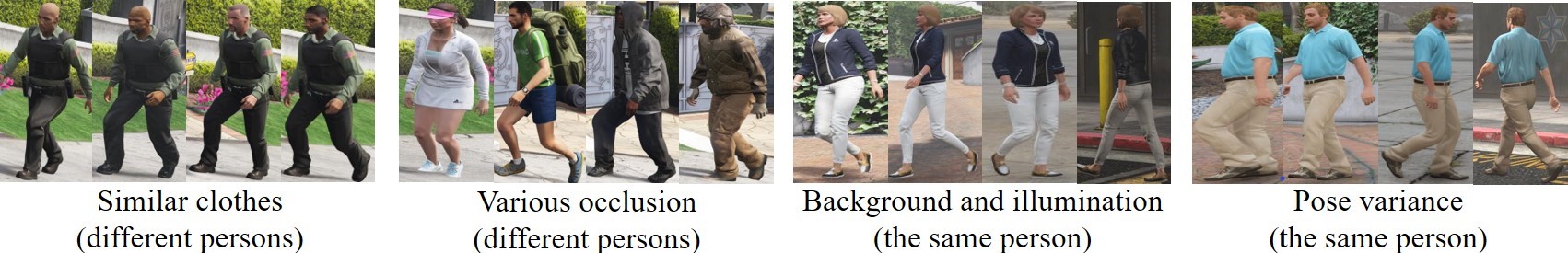}
\caption{\label{fig:identity}Examples of people with different clothes, various occlusions, background and illumination, as well as pose variance.}
\end{figure*}

\begin{figure*}[t]
\begin{centering}
\centering{}\includegraphics[width=1.7\columnwidth]{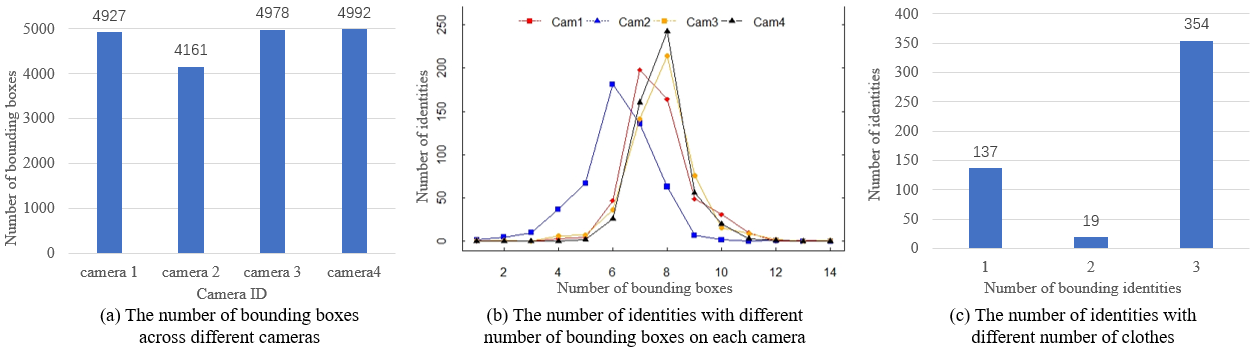}
\par\end{centering}
\caption{\label{fig:Statistical-data-of}Statistical data of VC-Clothes. We
show that the numbers of bounding boxes (i.e., person instances) from each camera are balanced (sub-figure (a)). The average
number of images per identity are between 6 to 8 for each camera (sub-figure (b)). Totally, 354 people
have 3 suits of clothes and 137 have only 1 suit of clothes (sub-figure (c)). All these
numbers demonstrate the research values of our VC-Clothes dataset on the clothes changeable ReID task.}
\end{figure*}

\section{Related Work}

\subsection{Clothes Inconsistency in ReID.}
There are few studies about \wan{the} clothes inconsistency problem in person \yang{ReID}. The biggest challenge is \wan{the} lack of sufficient applicable and well-annotated data with clothes inconsistency. RGB-D \cite{barbosa2012re} is the first \yang{applicable ReID dataset} proposed to solve this problem using additional depth information. However, \yang{the depth cue} is only applicable in indoor environments and the \yang{RGB-D} dataset is not sufficient to train a deep learning model (with only 79 identities and 1-2 clothes per identity). iQIYI-VID \cite{liu2018iqiyi} is a large scale multi-modal dataset collected from 600K video clips of 5,000 celebrities of various types. Although this dataset demonstrates great advantages in terms of size and variation in human poses, face quality, clothes, makeup, \textit{etc.}, the \yang{data are collected} from \yang{Internet} videos, instead of surveillance cameras. Another possible way is to automatically generate images with clothes differences using \yang{Generative Adversarial Networks}, \textit{e.g.} DG-Net \cite{zheng2019joint}. The encoders decompose each pedestrian image \yang{and project them} into two latent spaces: an appearance space and a structured space\yang{. By doing so}, one can change the color of clothes by switching the two latent space of different persons. However, this work did not dig into the importance of clothes-independent features\yang{, and the} appearance information \yang{still} contributes most \yang{to} the performance improvement. \yang{Moreover, it} can only change the color of clothes without altering human poses or illuminations at the same time. \yang{The recently introduced} PRCC \cite{yang2019person} dataset \yang{contains 33,698 real human} images of 221 different identities, \yang{which may be sufficient for research. However, it is only} collected in indoor laboratory environment \yang{against} simple backgrounds. Different from all the above \yang{works}, we propose a \yang{large} synthetic dataset which can \yang{better represent} the \yang{real demands}.

\subsection{Adopting Synthetic Data.}
Synthetic data is becoming increasingly popular in training deep learning models \cite{shrivastava2017learning}, because of the difficulty in building an ideal \yang{dataset in real environment} and \yang{the} requirements of huge human labor resources \yang{for data annotation}. Comparatively, synthetic data can be generated in an automated \yang{way}, therefore, promotes the development of many computer vision tasks like: semantic segmentation \cite{mccormac2017scenenet,ros2016synthia}, object detection \cite{peng2015learning,hattori2015learning}, pose estimation \cite{aubry2014seeing,busto2015adaptation,fabbri2018learning} \textit{et al}. In \yang{ReID}, only a few works \cite{bak2018domain,barbosa2018looking,sun2019dissecting} utilize synthetic data. \yang{Two of them} \cite{bak2018domain,sun2019dissecting} are designed to investigate illumination varieties or viewpoint diversities in \yang{ReID} instead of clothing changes. Although \yang{the other one} SOMAset \cite{barbosa2018looking} considers clothing changes, it does not discuss the influence of \yang{them} in depth. \yang{Differently, our two datasets contain} 512 identities with \yang{1{\usefont {T1}{qpl}{m}{n} \textasciitilde}3} different suits of clothes for each \yang{identity, and they can better} serve the research \yang{on} the clothes inconsistency \yang{issue in ReID}.

\section{New Benchmark Datasets}

Current person \yang{ReID} datasets are not sufficient for in-depth researches as they either \yang{lack} clothes inconsistency or \yang{limit in} data diversities \yang{such as missing outdoor scenarios}. Therefore, we decided to build new benchmark datasets. However, it is difficult to build a large-scale dataset of real surveillance data with significant clothes changes. Thus, we built a small one with 28 student volunteers for just preliminary research and put more efforts \yang{in} building a much larger dataset with realistic synthetic data generated by a powerful 3D video game engine. In this paper, we show that these two datasets are sufficient for supporting many valuable researches on the historically overlooked clothes inconsistency issue with our initial studies. Firstly, we give a brief \yang{introduction} to the two datasets, especially the synthetic one, which is unique and good for developing new robust \yang{ReID} models. We also prove its potential to serve as a new person \yang{ReID} benchmark for \yang{advancing the related researches}.

\subsection{The Real28 Dataset}

We name our real-scenario dataset ``Real28''. As can be seen in Fig. \ref{fig:real_samples} \yang{with the representative samples, the dataset} is collected in 3 different days (with different clothing) by 4 cameras\yang{. It consists of totally} 4,324 images \yang{from 28 different identities} with 2 indoor scenes and 2 outdoors. Since the size is not big enough for training deep learning models, we suggest only using this dataset for performance \yang{evaluation} and having the models trained on some other datasets, like the \yang{following} synthetic dataset VC-Clothes. We only split the whole dataset into \yang{a} query set and \yang{a} gallery set. We randomly select an image of each \yang{identity} in different scenes and different dates as the query image, while the other images \yang{are treated} as the gallery images. As a result, there are totally 336 images in the query set and 3,988 images in the gallery set.

\begin{table*}
\caption{\wan{Statistical} comparison between VC-Clothes and other popular \yang{ReID} datasets.}
\begin{centering}
\setlength{\tabcolsep}{2mm}{
\begin{tabular}{c|c|c|c|c|c|c}
\hline 
{Dataset}  & {DukeMTMC-reid}  & {Market1501}  & {CUHK03}  & {VC-Clothes}  & {CUHK01}  & {VIPeR}\tabularnewline
\hline 
\hline 
{\yang{No. of }BBoxes}  & {36,411}  & {32,668}  & {28,192}  & {19,060}  & {3,884}  & {1,264}\tabularnewline
\hline 
{\yang{No. of }Identities}  & {1,812}  & {1,501}  & {1,467}  & {512}  & {971}  & {632}\tabularnewline
\hline 
{\yang{No. of }Cameras}  & {8}  & {6}  & {2}  & {4}  & {10}  & {2}\tabularnewline
\hline 
{Detector}  & {hand}  & {DPM}  & {DPM, hand}  & {Mask \yang{RCNN}}  & {hand}  & {hand}\tabularnewline
\hline 
{Scene}  & {outdoor}  & {outdoor}  & {indoor}  & {indoor, outdoor}  & {indoor}  & {outdoor}\tabularnewline
\hline 
{Clothing change}  & {\ding{55}}  & {\ding{55}}  & {\ding{55}}  & {\ding{51}}  & {\ding{55}}  & {\ding{55}}\tabularnewline
\hline 
\end{tabular}}
\par\end{centering}
\label{tab:Dataset_Statistics} 
\end{table*}

\subsection{The VC-Clothes Dataset}

\subsubsection{Dataset construction}
\renewcommand{\thefootnote}{1}
The images are rendered by the Grand Theft Auto V\footnote{https://www.rockstargames.com/} (GTA5), which is a famous action-adventure game, with HD and realistic game graphics. Critically, it allows very convenient configurations of the clothes of each avatar/character and supports \yang{user-defined} environmental parameters, \yang{such as those for illumination, viewpoint} and background. \yang{Fabbri \emph{et al.} }\cite{fabbri2018learning} even made a pedestrian tracking and pose estimation dataset on top of it. GTA5 has a very large environment map, including thousands of realistic buildings, streets and spots. We select 4 different scenes -- 1 indoor scene and 3 outdoor scences\yang{, \emph{i.e.}, gate, street, natural scene and parking lot, virtually captured by 4 cameras (Cam1, Cam2, Cam3, Cam4), as shown in Fig.~\ref{fig:VC-Clothes_scene}}. We introduce additional illumination by changing not only the illumination of different scenes, but also the same scene \yang{under different time}, to make the images more realistic. In general, outdoor scenes have more lighting than indoors.

In the process of image generation, each person walks along a scheduled route, and cameras are \yang{set up and fixed at the chosen locations}. We have different camera views of front-left, front-right, and right-after, thus the \yang{human} faces may not be easily observed. However, each person still has very \yang{distinctive} biological traits, as demonstrated in Fig. \ref{fig:identity}. Thus, it is imperative to identify these persons by the features of \yang{face, gender, age, body figure, hairstyle} and so on, rather than the clothes. Notice we only change the clothes of persons, while maintaining their \yang{identities} unchanged. In our VC-Clothes dataset, each identity has \yang{1{\usefont {T1}{qpl}{m}{n} \textasciitilde}3 suits of clothes and the clothes of each identity are kept unchanged when captured by} the same camera. Furthermore, to facilitate experiments, we make all the identities wear the same clothes \yang{under} Cam2 and Cam3.

The main challenge of this dataset comes from the visual appearance of different clothes of the same identity. Other challenges also include different persons with similar clothes, various occlusion, illumination changes under different environments, and pose variance (see Fig. \ref{fig:identity})

\begin{table*}[ht]
\caption{The results of state-of-the-art methods on VC-Clothes and CUHK03 with
labeled bounding boxes. Note that $^{\ast}$ indicates that PCB adopts a new training/testing
protocol proposed in \cite{zhong2017re}, thus yields worse performance
than other methods.}
\begin{centering}
\setlength{\tabcolsep}{4.2mm}{
\begin{tabular}{c|c|c|c|c|c|c|c|c}
\hline 
\multirow{2}{*}{\small Methods} & \multicolumn{4}{c|}{{\small VC-Clothes}} & \multicolumn{4}{c}{\small CUHK03}\tabularnewline
\cline{2-9} 
 & \small{mAP}  & \small{R@1}  & \small{R@5}  & \small{R@10}  & \small{mAP}  & \small{R@1}  & \small{R@5}  & \small{R@10}\tabularnewline
\hline 
\small{LOMO+XQDA\cite{liao2015person}}  & \small{36.9 } & \small{42.3 } & \small{53.1 } & \small{57.7 } & \small{- } & \small{52.2 } & \small{82.2 } & \small{92.1}\tabularnewline
\small{GOG+XQDA\cite{matsukawa2016hierarchical}}  & \small{37.8 } & \small{42.9 } & \small{55.9 } & \small{61.8 } & \small{- } & \small{67.3 } & \small{91 } & \small{96}\tabularnewline
\hline 
\small{Baseline}  & \small{47.4 } & \small{50.1 } & \small{72.6 } & \small{80.2 } & \small{- } & \small{68.9 } & \small{79.7 } & \small{84.5}\tabularnewline
\small{MDLA\cite{qian2017multi}}  & \small{76.8 } & \small{88.9 } & \small{92.4 } & \small{93.9 } & \small{- } & \small{76.8 } & \small{96.1 } & \small{98.4}\tabularnewline
\small{PCB$^{\ast}$\cite{sun2018beyond}}  & \small{74.6 } & \small{87.7 } & \small{91.1 } & \small{92.8 } & \small{57.5 } & \small{63.7 } & \small{80.6 } & \small{86.9}\tabularnewline
\small{Part-aligned\cite{suh2018part}}  & \textbf{\small 79.7} & \textbf{\small 90.5} & \textbf{\small 94.3} & \textbf{\small 96.7} & {- } & \textbf{\small 88.0} & \textbf{\small 97.6} & \textbf{\small 98.6}\tabularnewline
\hline 
\end{tabular}}
\par\end{centering}
\label{tab:VC-Clothes_benchmark} 
\end{table*}

\subsubsection{Dataset properties}

VC-Clothes has 512 identities, 4 scenes (cameras) and on average 9 images/scene for each identity and \yang{a} total number of 19,060 images. Mask \yang{RCNN} \cite{he2017mask} is employed to detect \yang{persons and} automatically \yang{produce} bounding boxes. The detailed statistics of VC-Clothes are summarized in Table. \ref{tab:Dataset_Statistics}, in comparison with other widely used \yang{ReID} datasets, \yang{including} DukeMTMC-reid \cite{zheng2017unlabeled}, Market1501 \cite{zheng2015scalable}, CUHK03 \cite{li2014deepreid}, CUHK01\cite{li2012human} and VIPeR\cite{gray2008viewpoint}. More statistical analysis of this dataset can be seen in Fig. \ref{fig:Statistical-data-of}.

We equally split the dataset by identities\yang{:} 256 identities \yang{for} training and the other 256 \yang{for testing}. We randomly chose 4 images per person from each camera as query, \yang{and have} the other images serve as gallery images. \yang{Eventually}, we \yang{get totally 9,449} images \yang{in the} training, 1,020 images \yang{as queries and 8,591 others in} the gallery.

\subsubsection{The benchmark}

To make VC-Clothes dataset a qualified benchmark, we compare the results of some representative ReID methods on VC-Clothes with a representative benchmark, CUHK03.

\textbf{Problem definition.} We still follow the classical definition of person ReID task. Given a training dataset $D_{train}=\left\{ \mathbf{I}_{i},y_{i}\right\} {}_{i=1}^{N}$of $N$ images, where $\mathbf{I}_{i}$ and $y_{i}$ are the image data and its corresponding label, the goal of ReID is to learn an effective feature mapping $\Phi\left(\mathbf{I}\right)$ for the input image $\mathbf{I}$, such that images of the same identity have smaller distances in the feature space than those of different identities. In VC-Clothes, we target to find both the clothes consistency results and clothes inconsistency results at the same time.

\textbf{Evaluation metrics.} We use two main evaluation metrics: the Mean Average Precision (mAP) and the accuracy values at the first few indicative ranks, \emph{e.g.}, Rank-1 (in abbr. R@1), Rank-5 (R@5), and Rank-10 (R@10). MAP refers to the average accuracy rate of each relevant document retrieved by a query. The Rank-i accuracy means whether the first i query result contains the correct image.

\begin{table*}
\caption{Result differences between wearing the same clothes and changing clothes
(using two camera pairs as an example).}
\begin{centering}
{}%
\setlength{\tabcolsep}{3.5mm}{
\begin{tabular}{c|c|c|c|c|c|c|c|c}
\hline 
\multirow{3}{*}{Methods} & \multicolumn{4}{c|}{\textbf{Same Clothes}} & \multicolumn{4}{c}{\textbf{Change Clothes}}\tabularnewline
 & \multicolumn{4}{c|}{{ (1 suit/person, Cam2\&Cam3)}} & \multicolumn{4}{c}{{ (2 suits/person, Cam3\&Cam4)}}\tabularnewline
\cline{2-9} 
 & {mAP}  & {R@1}  & {R@5}  & {R@10}  & {mAP}  & {R@1}  & {R@5}  & {R@10}\tabularnewline
\hline 
{LOMO+XQDA \cite{liao2015person}}  & {83.3 } & {86.2 } & {93.8 } & {95.1 } & {30.9 } & {34.5 } & {44.2 } & {49.7}\tabularnewline
{GOG+XQDA \cite{matsukawa2016hierarchical}}  & {86.7 } & {90.5}  & {94.5}  & {95.7}  & {31.3 } & { 35.7 }  & { 48.3}  & { 54.2}\tabularnewline
\hline 
{Baseline}  & {74.3}  & {79.3}  & {92.8}  & {94.6}  & {32.4 } & {36.4}  & {57.4}  & {65.0}\tabularnewline
{MDLA \cite{qian2017multi}}  & {93.9}  & {94.3}  & {95.7}  & {96.3}  & {60.8}  & {59.2}  & {67.3}  & {73.5}\tabularnewline
{PCB \cite{sun2018beyond}}  & \textbf{94.3} & \textbf{94.7} & \textbf{96.3} & \textbf{96.7} & {62.2}  & {62.0}  & {68.2}  & {72.9}\tabularnewline
{Part-aligned \cite{suh2018part}}  & {93.4}  & {93.9}  & {96.0}  & {96.7}  & \textbf{67.3} & \textbf{69.4} & \textbf{77.5} & \textbf{84.5}\tabularnewline
\hline 
\end{tabular}}
\par\end{centering}
\label{tab:Harder_problem} 
\end{table*}

\begin{table*}
\caption{Generalization to unseen clothes (training with 2 suits/person using
Cam3\&Cam4).}
\begin{centering}
{}%
\setlength{\tabcolsep}{3.9mm}{
\begin{tabular}{c|c|c|c|c|c|c|c|c}
\hline 
\multirow{3}{*}{Methods} & \multicolumn{4}{c|}{{Query in }\textbf{seen}{ suits (Cam2)}} & \multicolumn{4}{c}{{Query in }\textbf{unseen}{ suits (Cam1)}}\tabularnewline
 & \multicolumn{4}{c|}{{Gallery (Cam3\&Cam4)}} & \multicolumn{4}{c}{{Gallery (Cam3\&Cam4)}}\tabularnewline
\cline{2-9} 
 & {mAP}  & {R@1}  & {R@5}  & {R@10}  & {mAP}  & {R@1}  & {R@5}  & {R@10}\tabularnewline
\hline 
{Baseline}  & {28.8} & {27.5} & {53.1} & {63.9} & {19.5} & {18.4} & {37.5} & {49.6}\tabularnewline
{MDLA\cite{qian2017multi}}  & {59.0} & {57.7} & {75.2} & {81.1} & {43.2} & {40.5} & {63.5} & {73.5}\tabularnewline
{PCB\cite{sun2018beyond}}  & \textbf{73.9} & \textbf{72.3} & {83.8} & {87.8} & {55.6} & {53.9} & {71.2} & {77.4}\tabularnewline
{Part-aligned\cite{suh2018part}}  & {70.5} & {68.9} & \textbf{84.7} & \textbf{89.3} & \textbf{57.9} & \textbf{55.6} & \textbf{78.8} & \textbf{85.8}\tabularnewline
\hline 
\end{tabular}}
\par\end{centering}
\label{tab:Unseen_Clothes} 
\end{table*}

\textbf{Competitors.} We compare several  hand-crafted methods, including LOMO+XQDA \cite{liao2015person}, GOG+XQDA \cite{matsukawa2016hierarchical}; DNN methods: MDLA \cite{qian2017multi}, PCB \cite{sun2018beyond}, Part-aligned \cite{suh2018part}. We use the original setup of the 5 state-of-the-art methods in our experiments. In addition, we also provide a baseline model using the widely adopted ResNet50 \cite{he2016deep} network pre-trained on the ImageNet \cite{krizhevsky2012imagenet} dataset. During training, the batch-size is set to 128 and the model is trained for 200 epochs with an initial learning rate of 0.0001, which will decay by a factor of 0.01 every 50 epochs.

\textbf{Benchmark results.} As shown in Table \ref{tab:VC-Clothes_benchmark}, the performances trend of different methods on the VC-Clothes dataset is similar as on CUHK03, showing that VC-Clothes can reflect real-world performance. It also enables different design of sub settings, thus we can systematically study the influence of clothes-inconsistency problem. All of the above factors make VC-Clothes a rational benchmark for person ReID model evaluation.

\section{Influence of Clothing Change\label{sec:Influence}}

When designing experimental settings, we mainly consider whether people change their clothes or not, and have all identities maintaining their clothes over two cameras (Cam2 and Cam3) and changing their clothes over any other camera pairs. Such a setting enables \wan{the} study of how can clothes inconsistency \wan{influences} the performance of ReID.

\subsection{Making ReID a Much Harder Problem}

\label{subsec:Clothes-bias-in}

We first conduct a typical experiment on justifying the performance of ReID on two indicative settings: \textbf{Same Clothes} (Cam2 and Cam3) vs. \textbf{Change Clothes} (Cam3 and Cam4), using all the representative state-of-the-art methods. \wan{We intentionally select the images from specific cameras in training and testing set to serve as the data for change clothes and same clothes settings.} As can be seen in Table \ref{tab:Harder_problem}, clothes inconsistency brings about 30\% to 50\% performance drop to DNN-based methods, and even more to hand-crafted methods, in terms of mAP and Rank-1 accuracy. Although camera diversities may also influence the results to some extent, the primary cause of the dramatic performance decrease should be the changing of clothes.

This indicates that clothes inconsistency is essentially a very subtle problem, which unfortunately has been overlooked in previous works. This is easy to understand as clothes usually cover most of the body, which makes it hard to extract identity-sensitive but clothes-independent appearance features. Note that the performance drop of the Part-aligned method \cite{suh2018part} is the least in all the compared methods, indicating finer and more detailed modeling of appearance is needed for handling clothing changes.

\subsection{Changing Learned Representations}

\begin{figure}
\centering{}\includegraphics[width=0.35\textwidth]{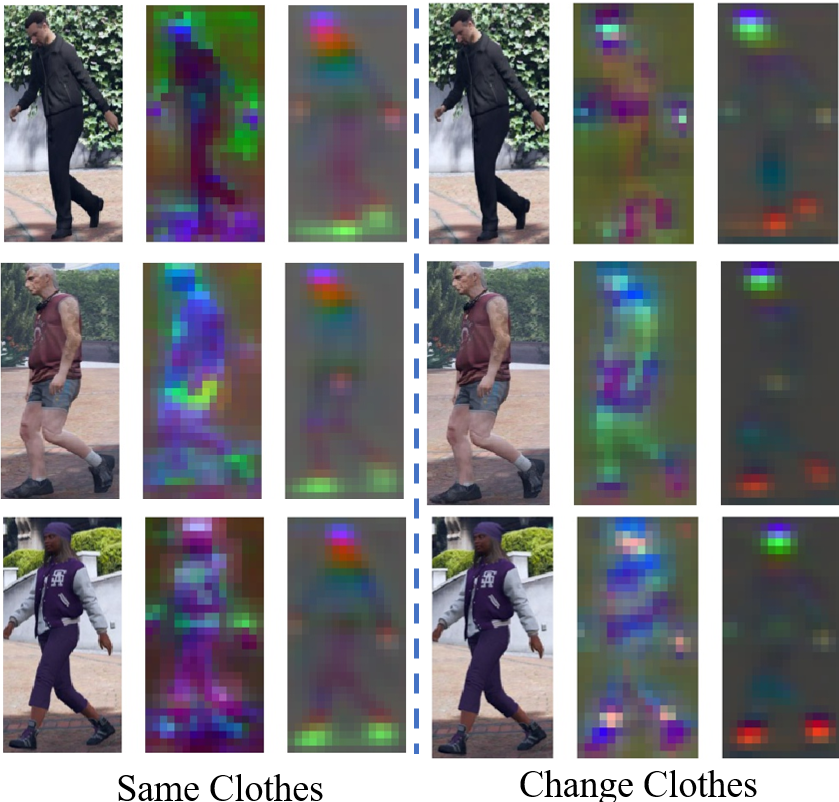}\caption{\label{fig:The-feature-maps}The feature maps of Part-Aligned method
trained on ``Same Clothes'' setting (Cam2\&Cam3) on the left and
``Change Clothes'' setting (Cam1\&Cam2) on the right.}
\end{figure}

In Fig. \ref{fig:The-feature-maps}, we give a visualization of the appearance maps and part maps of the Part-aligned model \cite{suh2018part} under the ``Same Clothes'' setting (Cam2 and Cam3) and ``Change Clothes'' (Cam1 and Cam2) settings (seeing Tab.~\ref{tab:Harder_problem}). For a given input image (left), the global appearance (center) and body parts (right) are represented using different colors. The same color implies similar features and brighter locations denote more significant body parts.

In the ``Same Clothes'' setting where there is no clothes inconsistency, the appearance maps of body parts are much brighter than that in the ``Change Clothes'' setting and the colors are more scattered. Obvious attention on faces and feet can be seen in the part map of the ``Change Clothes'' setting, which means learning identity-sensitive and clothes-independent representations require exploration of the details of smaller areas instead of the large body parts.

\subsection{Generalization to Unseen Clothes}

To evaluate the ReID models' ability in handling clothes inconsistency of the same identities, training using the test data from the same camera pair (as shown in Tab. \ref{tab:Harder_problem}) is not enough, because the learned models may overfit specific camera viewpoints. Therefore, we propose to test by querying with data from another new camera.

In the VC-Clothes dataset, each camera corresponds to a type of suit for each identity. We can train models on Cam3 and Cam4, and test on two different cameras: Cam2 and Cam1. The two cameras represent two different settings. Choosing Cam2 means querying with human instances in \textbf{seen} suits, as the clothes of each person are the same for Cam2 and Cam3. Differently, Cam1 records a new \textbf{unseen} suit for each identity. The two models are trained on the same setting but test on two different settings, thus making it possible to evaluate not only the performance on querying with a new camera, but also the generalization ability on handling unseen clothes, which is believed to be more important when clothing changes do happen.

The experimental results are presented in Table \ref{tab:Unseen_Clothes}. Since DNN-based methods perform much better than hand-crafted ones, we only list the results of DNN-based methods. As the result shows, querying with a new camera is not a big issue when people keep their clothes unchanged, but wearing unseen suits do bring significant extra difficulties and limit the generalization ability of the models.

\begin{table}
\begin{centering}
 \caption{\label{tab:clothes_independence}Different combination of models pre-trained and tested on ``All
clothes'' or ``Same clothes'' settings.}
\setlength{\tabcolsep}{1.7mm}{
\begin{tabular}{c|c|c|c|c|c|c}
\hline 
\multirow{3}{*}{\small{Pre-trained}} & \multicolumn{6}{c}{\small{Test}}\tabularnewline
\cline{2-7} 
 & \multicolumn{3}{c|}{\small{All clothes}} & \multicolumn{3}{c}{\small{Same clothes}}\tabularnewline
\cline{2-7} 
 & \small{R@1} & \small{R@10} & \small{mAP} & \small{R@1} & \small{R@10} & \small{mAP}\tabularnewline
\hline 
\small{All clothes} & \small{90.8} & \small{96.7} & \small{82.9} & \small{92.9} & \small{96.1} & \small{92.4}\tabularnewline
\small{Same clothes} & \small{63.1} & \small{69.7} & \small{43.1} & \small{93.9} & \small{96.7} & \small{93.4}\tabularnewline
\hline 
\end{tabular}}
\par\end{centering}
\end{table}

\subsection{Benefits of Training with Clothes Inconsistency}

To prove the advantage of models trained on \wan{the} dataset with clothes inconsistency, we test the generalization ability of model pre-trained on ``\textbf{All Clothes}'' setting (images from all cameras) and ``\textbf{Same Clothes}'' setting (images for only Cam2 and Cam3) and make a confusion matrix of the two models tested on ``All Clothes'' setting and ``Same Clothes'' setting, respectively. Specifically, the ``Same Clothes'' setting imitates the general person ReID dataset, while the ``All Clothes'' setting is actually VC-Clothes dataset.

As shown in Table \ref{tab:clothes_independence}, the performance drop of model pre-trained on ``All Clothes'' setting and tested on ``Same Clothes'' setting is ignorable, about only 1\% on VC-Clothes (in terms of R@1 and mAP, while R@10 has fewer differences), compared with the model pre-trained and tested both on ``Same Clothes'' setting. However, a dramatic decrease of performance appears when the pre-trained ``Same Clothes'' model is tested on the ``All Clothes'' setting (Rank-1 and mAP reduce by 27.7\% and 39.8\% respectively). Comparatively, models trained on the dataset with clothes inconsistency can perform well on both clothes consistency setting and clothes inconsistency setting, while models trained only on the dataset with only clothes consistency, does not show good generalization ability on clothes inconsistency setting.

\begin{table*}[t]
\caption{Improving the performance by enriching the training data.\label{tab:Enrich_data}}
\begin{centering}
 {}%
 \setlength{\tabcolsep}{0.5mm}{
\begin{tabular}{c|c|c|c|c|c|c|c|c|c|c|c|c}
\hline 
\multirow{3}{*}{\small{Methods}} & \multicolumn{4}{c|}{\small{{Train: 2 suits/person (Cam3\&Cam4)}}} & \multicolumn{4}{c|}{\small{{Train: }add 1 \textbf{unseen}{
suit/person (add Cam1)}}} & \multicolumn{4}{c}{\small{{ Train: }add 1 \textbf{seen}{
suit/person (add Cam2)}}}\tabularnewline
 & \multicolumn{4}{c|}{\small{{ Test: 2 suits/person (Cam3\&Cam4) }}} & \multicolumn{4}{c|}{\small{{ Test: 2 suits/person (Cam3\&Cam4) }}} & \multicolumn{4}{c}{\small{{ Test: 2 suits/person (Cam3\&Cam4) }}}\tabularnewline
\cline{2-13} 
 & \small{{mAP}} & \small{{R@1}} & \small{{R@5}} & \small{{R@10}} & \small{{mAP}} & \small{{R@1}} & \small{{R@5}} & \small{{R@10}} & \small{{mAP}} & \small{{R@1}} & \small{{R@5} } & \small{{R@10} }\tabularnewline
\hline 
\small{{Baseline}} & \small{32.4} & \small{36.4} & \small{57.4} & \small{65.0} & \small{34.4} & \small{38.8} & \small{63.9} & \small{73.5} & \small{35.0} & \small{36.7} & \small{47.6} & \small{52.5}\tabularnewline
\small{{MDLA}} & \small{60.8} & \small{59.2} & \small{67.3} & \small{73.5} & \small{62.6} & \small{63.5} & \small{73.1} & \small{79.2} & \small{54.2} & \small{55.1} & \small{61.6} & \small{68.2}\tabularnewline
\small{{ PCB} } & \small{62.2} & \small{62.0} & \small{68.2} & \small{72.9} & \small{66.9} & \small{66.3} & \small{73.3} & \small{79.2} & \small{58.1} & \small{57.5} & \small{64.7} & \small{70.6}\tabularnewline
\small{{ Part-aligned} } & \textbf{\small67.3} & \textbf{\small69.4} & \textbf{\small77.5} & \textbf{\small84.5} & \textbf{\small68.3} & \textbf{\small69.4} & \textbf{\small79.0} & \textbf{\small85.9} & \textbf{\small63.9} & \textbf{\small64.1} & \textbf{\small73.7} & \textbf{80.0}\tabularnewline
\hline 
\end{tabular}}
\par\end{centering}
{}
\vspace{-0.1in}
\end{table*}

\section{Preliminary Solutions}
In this section, we further discuss some preliminary solutions towards the robustness against clothes inconsistency. In particular, we present several straightforward attempts. \wan{We hope} our experimental results \wan{can} motivate and inspire further studies in such an important issue.

\subsection{Enriching Training Data}

\begin{figure}[t]
\begin{centering}
\centering{}\includegraphics[scale=0.3]{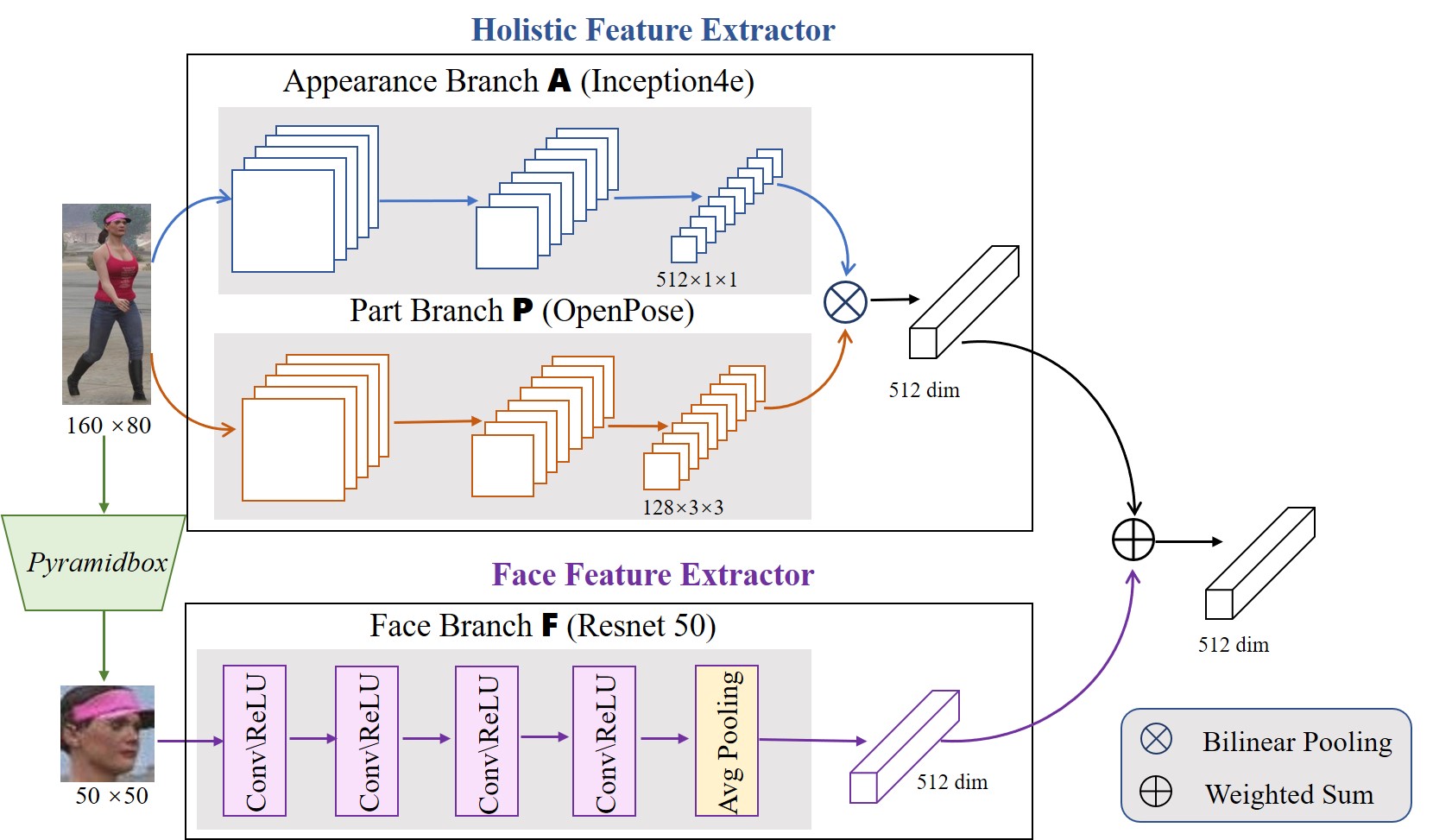}
\par\end{centering}
\caption{\label{fig:Overview-of-our}Overview of our model. The two 512 dimension feature vectors from holistic feature extractor and face feature extractor are concatenated using weighted sum.}
\end{figure}

One common strategy to improve the performance of DNN-based models is providing more training data. We treat the models trained on Cam3 and Cam4 as the basic settings, and add data from one more camera to enlarge the training set. By choosing either Cam1 or Cam2 as \wan{an} additional camera, we can get two different settings. Adding Cam1 means adding one more unseen suit/person, while adding Cam2 means adding a seen suit/person, as Cam2 and Cam3 share the same clothes for each person.

The experimental results of these three different settings are presented in Table \ref{tab:Enrich_data}. Adding 1 unseen suit/person significantly boosts the performance of all methods, while the opposite effect can be observed when adding 1 seen suit/person. It may indicate that adding unseen suits can make the DNN-based models work harder on generating more clothes-independent representations, which is more effective for ReID with clothing changes. Therefore, adding more training data is proved to be effective when introducing new clothes.

\subsection{Exploring Exposed Body Parts: the Faces}


\textbf{Overview of the fusion model.} In order to improve the performance
of current ReID models on top of VC-Clothes, we propose a 3-stream
Appearance, Part and Face Extractor Network (3APF). As summarized
in Fig. \ref{fig:Overview-of-our}, it is composed of two main components:
holistic feature extractor (ReID part) and a local face feature extractor
(face part).

\textbf{Holistic feature extractor.} We directly utilize the Part-aligned
network $f_{1}\left(\cdot\right)$ \cite{suh2018part} as the holistic
feature extractor following the original setting. In particular, this
network learns the appearance $A\left(\cdot\right)$ and body part
$P\left(\cdot\right)$ branches, which are further combined by a bi-linear
pooling. Thus, this network computes the features as in Eq. \ref{eq1}.

\begin{equation}
f_{1}\left(\mathbf{I}\right)=\mathrm{pooling}\left(A\left(\mathbf{I}\right)\otimes P\left(\mathbf{I}\right)\right)\label{eq1}
\end{equation}

\textbf{Local face feature extractor.} We utilize Pyramidbox \cite{tang2018pyramidbox} for face detection; and only 80\% person images have detectable faces, because of non-frontal faces. The confidence of the face bounding boxes is set to 0.8. Further, we expand the detected bounding boxes of faces by 15 pixels in four directions (up, down, left and right, respectively) to cover more facial area. All face images are then resized to the same size of $50\times50$ pixels. As for the images with undetectable faces, we use only the feature from a holistic feature extractor for identification.

We use the widely adopted Resnet50 network $F\left(\cdot\right)$ \cite{he2016deep} pre-trained on MS-1M \cite{guo2016msceleb} dataset to extract face features $f_{2}\left(\mathbf{I}\right)=F\left(G\left(\mathbf{I}\right)\right)$, where $G$ is a face detector. The final feature vector is the weighted mean result of two features as in Eq. \ref{eq2}, where $w$ is the hyper-parameter to determine the proportion of each part.

\begin{equation}
f=w\cdot f_{1}+\left(1-w\right)\cdot f_{2}\label{eq2}
\end{equation}

\textbf{Improvement brought by faces}
The visualization of Fig. \ref{fig:The-feature-maps} demonstrates the significance of face, when there exists clothes inconsistency. To describe the influence of face to the ReID part in a quantitative way, we employ the 3APF model to control the proportion of face and ReID part by adjusting the weight. The larger the weight, the more important of ReID part in our model. As can be seen in Table. \ref{tab:3APF}, when we adjust the weight to a certain value (0.95), a significant improvement on mAP (82.1\%) can be witnessed over single face extractor and holistic extractor.

\subsection{Values of VC-Clothes to Real Dataset}

To evaluate whether our pre-trained models on VC-Clothes (synthetic dataset with clothes inconsistency) has some real-world applications and prove its advantages over Market1501 (real dataset without clothes inconsistency). We adopt 2 different ways to transfer the model pre-trained on VC-Clothes to real dataset Real28. One is directly applying the 3APF models trained on VC-Clothes or Market1501 to Real28 or RGB-D. Another way is using finetuning to learn better representations of real data with clothes inconsistency. ``VC-Clothes{\scriptsize{}{}+FT}'' means pre-trained on VC-Clothes and fine-tuned on Market1501, while ``Market1501{\scriptsize{}{}+FT}'' denotes the opposite.

As can be seen in Table. \ref{tab:pretrained-model}, directly applying the model trained on VC-Clothes is hard to get good results, but we can get much better results (VC-Clothes{\scriptsize{}{}+FT}) by simply fine-tuning it on a third real dataset. Moreover, our synthetic data can also help model pre-trained on real datasets to generate better performance on the tasks with clothes inconsistency (Market1501{\scriptsize{}{}+FT}). This phenomenon is especially significant on ``Real28'', where more clothes changes exist.

\begin{table}
\begin{centering}
{\caption{The results of our 3APF model on VC-Clothes.\label{tab:3APF}}
}
\setlength{\tabcolsep}{3mm}{
\begin{tabular}{c|c|c|c|c}
\hline 
\small{Model }  & \small{mAP }  & \small{R@1 }  & \small{R@5 }  & \small{R@10}\tabularnewline
\hline 
\small{Face extractor}  & \small{60.8}  & \small{63.5}  & \small{87.0}  & \small{91.4}\tabularnewline
\small{Holistic extractor }  & \small{79.7 }  & \textbf{\small 90.5 }  & \small{94.3 }  & \small{96.7}\tabularnewline
\small{Ours}  & \textbf{\small 82.1}  & \small {90.2} & \textbf{\small 94.7}  & \textbf{\small 96.8}\tabularnewline
\hline 
\end{tabular}}
\par\end{centering}
\end{table}

\begin{table}
\begin{centering}
 \caption{Comparison of three different ways to transfer models to real test dataset with clothes inconsistency. \label{tab:pretrained-model}}
\setlength{\tabcolsep}{1.4mm}{
\begin{tabular}{c|c|c|c|c|c|c}
\hline 
\multirow{3}{*}{\small Method} & \multicolumn{4}{c|}{\small Real28} & \multicolumn{2}{c}{{\small RGB-D}}\tabularnewline
\cline{2-7} 
  & \multicolumn{2}{c|}{{\small Different clothes}} & \multicolumn{2}{c|}{\small{All Clothes}} & \multicolumn{2}{c}{\small{All clothes}}\tabularnewline
\cline{2-7} 
  & \small{R@1}  & \small{mAP}  & \small{R@1}  & \small{mAP}  & \small{R@1}  & \small{mAP}\tabularnewline
\hline 
\small{VC-Clothes}  & \small{5.2}  & \small{3.6}  & \small{22.7}  & \small{8.2}  & \small{17.2}  & \small{9.4}\tabularnewline
\small{Market1501}  & \small{8.2}  & \small{6.1}  & \small{48.1}  & \small{19.1}  & \small{27.9}  & \small{20.0}\tabularnewline
\small{VC-Clothes}{\scriptsize{}{}+FT}  & \small{11.0}  & \textbf{\small 10.6} & {\textbf{\small 55.3}}  & {\textbf{\small 24.3}}  & \small{26.6}  & \small{16.0}\tabularnewline
\small{Market1501}{\scriptsize{}{}+FT}  & {\textbf{\small 12.4}}  & \small{7.6}  & \small{54.0}  & \small{21.8}  & \textbf{\small 27.9} & \textbf{\small 20.0}\tabularnewline
\hline 
\end{tabular}}
\par\end{centering}
\end{table}

\section{Conclusions and Discussions}
In this paper, we studied the clothes inconsistency problem in person ReID. Such a setting is much realistic and helpful in tackling \wan{the} longtime person ReID task. Because of the huge difficulty in collecting a large-scale dataset in real scenes, we just collect a small one for testing. Alternatively, we proposed to use the GTA5 game engine to render a large-scale synthetic dataset ``VC-Clothes'' with desired properties. With these two benchmark datasets, we conducted a series of pilot studies, which helps us better understand the influence and importance of this new problem. Finally, we also investigated some straightforward solutions for improving the performance on VC-Clothes and ways to transfer the model to \wan{the} real test dataset. It is hopefully to motivate and inspire further studies \wan{on} this important issue.

\noindent \textbf{Acknowledgement.} This work was supported in part by Microsoft Research Asia Collaborative Research Grant.

{\small
\bibliographystyle{ieee_fullname}
\bibliography{sample-base}
}

\end{document}